\documentclass[accepted]{uai2021} 

\usepackage[british]{babel}

\usepackage{natbib} 
\usepackage{mathtools} 
\usepackage{booktabs} 
\usepackage{tikz} 

\newcount\Comments  
\definecolor{darkgreen}{rgb}{0,0.5,0}
\definecolor{purple}{rgb}{1,0,1}
\newcommand{\kibitz}[2]{\ifnum\Comments=0\textcolor{#1}{#2}\fi}

\usepackage{mathtools}
\usepackage{amsmath}
\usepackage{amssymb}
\usepackage{amsbsy}
\newtheorem{definition}{Definition}
\usepackage{comment}
\usepackage{multirow}
\newtheorem{remark}{Remark}
\DeclareMathOperator*{\argmax}{arg\,max}

\usepackage[absolute,overlay]{textpos}

\title{BayLIME: Bayesian Local Interpretable Model-Agnostic Explanations}

\author[1,2]{\href{mailto:Xingyu Zhao <xingyu.zhao@liverpool.ac.uk>?Subject=Your UAI 2021 paper}{Xingyu Zhao}{}} 
\author[1]{Wei Huang}
\author[1]{Xiaowei Huang}
\author[2,3,4]{Valentin Robu}
\author[2]{David Flynn}
\affil[1]{%
Department of Computer Science, University of Liverpool, Liverpool, L69 3BX, U.K.
}
\affil[2]{%
School of Engineering \& Physical Sciences, Heriot-Watt University, Edinburgh, EH14 4AS, U.K.
}
\affil[3]{%
Centrum voor Wiskunde en Informatica, Science Park 123, 1098 XG Amsterdam, The Netherlands
}

\affil[4]{Delft University of Technology, Algorithmics Group, EEMCS, 2628 XE Delft, The Netherlands}

\begin{document}
\begin{textblock*}{20cm}(1cm,1cm)
\textcolor{red}{Preprint accepted by UAI2021. To appear in the UAI2021 volume of Proceedings of Machine Learning Research (PMLR)}.
\end{textblock*}
\maketitle

\begin{abstract}
Given the pressing need for assuring algorithmic transparency, Explainable AI (XAI) has emerged as one of the key areas of AI research. In this paper, we develop a novel Bayesian extension to the LIME framework, one of the most widely used approaches in XAI -- which we call BayLIME. Compared to LIME, BayLIME exploits prior knowledge and Bayesian reasoning to improve both the consistency in repeated explanations of a single prediction and the robustness to kernel settings. BayLIME also exhibits better explanation fidelity than the state-of-the-art (LIME, SHAP and GradCAM) by its ability to integrate prior knowledge from, e.g., a variety of other XAI techniques, as well as verification and validation (V\&V) methods. We demonstrate the desirable properties of BayLIME through both theoretical analysis and extensive experiments.
\end{abstract}

\section{Introduction}
\label{sec_introduction}

A key challenge to wide adoption of AI methods is their perceived lack of transparency. The black-box nature of many AI methods means they do not provide human users with direct explanations of their predictions. This led to growing interest in Explainable AI (XAI) -- a research field that aims at improving the trust and transparency of AI. Explainable AI (XAI) methods can be classified by various criteria \citep[Chpt.~2.2]{molnar_interpretable_2020}, such as model-specific vs model-agnostic or local (instance-wise) vs global (entire model). 
Readers are referred to \citep{huang_survey_2020,adadi_peeking_2018} for a comprehensive review.

In this work, we focus on the class of XAI methods using local surrogate models for explaining individual predictions -- specifically, we develop a novel Bayesian extension to the most popular method in this category: Local Interpretable Model-agnostic Explanations (LIME) \citep{ribeiro_why_2016}. 
Despite its very considerable success in both research and practice, LIME has several weaknesses, the most significant of which are the lack of \emph{consistency in repeated explanations of a single prediction} and \emph{robustness to kernel settings}. Meanwhile, higher \textit{explanation fidelity} is also expected in many settings. 
Arguably, these three properties are among the most desirable for an XAI method to have. 

\begin{figure*}[ht]
	\centering
	\includegraphics[width=1\linewidth]{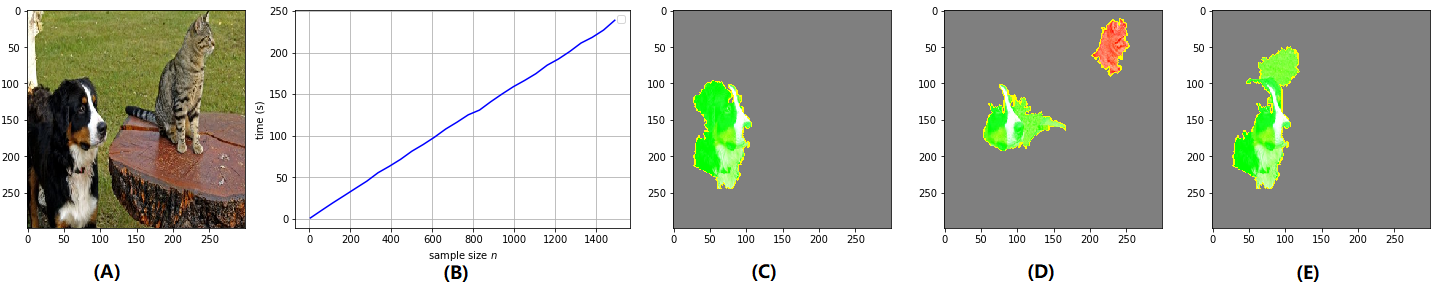}
	\caption{(A) The instance with prediction ``Bernese mountain dog'' under explanation; (B) LIME computational time as a function of perturbed sample size $n$; (C)-(E) Three repeated and noticeably \textit{inconsistent} LIME explanations when $n=100$, showing the top 4 features contributing toward and against the prediction (shaded green and red respectively).}
	\label{fig_lime_time_inconsist}
\end{figure*}

The inconsistency of LIME, where different explanations can be generated for the same prediction, has been identified as a critical issue by several prior works
~\citep{zafer_dlime_2019,shankara_alime_2019}. This is caused by the randomness in generating perturbed samples (of the instance under explanation) that are used for the training of local surrogate models. 
Plainly stated, a smaller sample size leads to greater uncertainty of such randomness. This inconsistency has been argued to limit its usefulness in critical applications such as the medical domain, where consistency is highly required~\citep{zafer_dlime_2019}.
 As an example, in Fig.~\ref{fig_lime_time_inconsist}, the pretrained Convolutional Neural Network (CNN) InceptionV3 \citep{szegedy_rethinking_2016} predicts the instance of Fig.~\ref{fig_lime_time_inconsist} (A) as ``Bernese mountain dog'' (top-1 label).
To explain this prediction, we vary the size of perturbed samples (denoted as $n$) and record the time consumption in Fig.~\ref{fig_lime_time_inconsist} (B). We see that the computational time is \emph{linear} with respect to $n$. If, say, an application requires LIME to respond in 20s, we have to limit $n$ to around 100 (in our case). Then, we may easily get three \textit{inconsistent} explanations, as shown in Fig.~\ref{fig_lime_time_inconsist} (C)-(E), in three repeated runs of LIME. 





The other problem of LIME motivating this work relates to defining the right \textit{locality} -- the ``neighbourhood''  on which a local surrogate is trained to approximate the AI model \citep{laugel_defining_2018}. As noted by \citep[Chpt.~5.7]{molnar_interpretable_2020} as a ``very big and unsolved problem'', how to choose a kernel setting in LIME to properly define a neighbourhood is challenging. Even worse, explanations can be easily turned around by changing kernel settings (cf. the example in \citep[Chpt.~5.7]{molnar_interpretable_2020}). Thus, an enhancement to LIME so that it is robust to the kernel settings is called for.

The goal of XAI is to provide good explanations. Despite the various definitions on the \textit{goodness} of explanations, fidelity -- how truthfully the explanation represents the underlying AI decision
-- is the most important criterion. Any new XAI method enhancing other properties should preserve the fidelity (if not improving), compared to the state-of-the-art.

In this paper, we propose a novel enhancement of LIME, named as BayLIME, by utilising a \emph{Bayesian} local surrogate model, which we show analytically is a ``Bayesian principled weighted sum'' of the prior knowledge and the estimates based on new samples (as LIME). The weights consist of parameters with dedicated meanings that can be either automatically fitted from samples (Bayesian model selection) or elicited from application-specific prior knowledge (e.g., V\&V methods). Our experiments show that BayLIME significantly improves over LIME with respect to a few objective measures, because the prior knowledge is independent from the causes of aforementioned inconsistency and un-robustness (thus benefits both properties) and also includes additional useful information  that improves the fidelity.

The contributions of our work include: 

(1) Developing a new XAI method, BayLIME, that provides a Bayesian principled mechanism for the \textit{embedding} of prior knowledge; 

(2) Introducing and designing 
quantitative metrics on both the consistency in repeated explanations and robustness to kernel settings that are generic to all feature-ranking based XAI techniques; 

(3) Three typical ways of obtaining prior knowledge and eliciting prior parameters, including explanations on similar instances, other diverse XAI methods and V\&V methods, to illustrate potential use cases of BayLIME; 

(4) An extensive set of experiments, exhibiting the superior performance of BayLIME over LIME on aforementioned
consistency and robustness, and higher fidelity over the state-of-the-art (LIME, SHAP, GradCAM); 

(5) A prototype tool with tutorials/experimental-results.


\section{Consistency in Repeated Explanations}
\label{sec_consistency}

LIME is known to be unstable in terms of 
generating inconsistent explanations in repeated runs 
\citep{zafer_dlime_2019,shankara_alime_2019}.
This is 
due to the randomness in the 
perturbed samples used 
to train the local surrogate model. 
%
While randomness can be reduced by enlarging the sample size, it might be impractical in real-world applications, particularly those applications with strict efficiency constraints (e.g., low response time).


It has been noted by the original LIME paper \citep{ribeiro_why_2016} that the time required to produce an explanation is \textit{dominated} by the complexity of the black-box AI model. As shown in Fig.~\ref{fig_lime_time_inconsist} (B), to improve LIME's efficiency, we cannot effectively tune other LIME arguments,
rather the best option is to limit the \textit{number of queries made} to the AI model (which, indeed, is very costly when the AI model is deep \citep{he_convolutional_2015}) when generating labels for the $n$ samples (step 3 of the LIME workflow in Appendix \ref{sec_app_preliminaries}). However, this solution leads to a severer inconsistency issue.


To compare between BayLIME and LIME on their inconsistency, we take Kendall's~W \citep{kendall_problem_1939} to measure the agreement among raters (i.e. repeated explanations in our case), which ranges from 0 (no agreement) to 1 (complete agreement). However, Kendall's~W only considers the discrete ranks of features, 
and cannot discriminate explanations with the same ranking of features but different importance vectors. To complement Kendall's~W in such corner cases, we also introduce a new metric based on the \textit{index of dispersion} (IoD) of each feature in repeated runs. The new metric weights the IoD of the rank of each feature by its importance, cf. Appendix~\ref{sec_app_incon_measure} for a formal definition.


\section{Robustness to Kernel Settings}
\label{sec_robust}

Another notorious problem of LIME (or any approach with a notion of localisation) is how to meaningfully define a ``neighbourhood'' of the instance of interest (as required by the step 4 of LIME depicted in Appendix~\ref{sec_app_preliminaries}). 
By default, LIME uses an exponential smoothing kernel to define the neighbourhood with a given kernel width
that determines how large the neighbourhood is.
While intuitively a smaller kernel width means that an instance needs to be closer to influence the local model (vice versa), no effective way exists to find the best kernel settings. The best strategy for now is to try different kernel settings and see if the explanations make sense, which inevitably is subject to errors/bias. Moreover, 
an explanation may be easily turned around by changing the kernel settings \citep[Chpt.~5.7]{molnar_interpretable_2020}.

To compare between BayLIME and LIME, we explore the robustness to the \textit{kernel width} parameter (denoted by $l$ later) in LIME's default exponential kernel function. Other kernel setting parameters\footnote{E.g., the distance parameter in a periodic kernel.} can be studied in a similar way.
\begin{definition}
	To explain a given instance $i$, we denote $h_i: (0,+\infty) \rightarrow \mathbb{R}^m $ as the importance vector of 
	$m$ features taking $l$ as the kernel width setting. For any pair of kernel width parameters $l_1$ and $l_2$, there exists a global Lipschitz value $\mathcal{L} \in \mathbb{R}$ such that $||h_i(l_1)-h_i(l_2)||\leq \mathcal{L} ||l_1-l_2||$.
\end{definition}
Intuitively, this global Lipschitz value quantifies the robustness of an explainer to the 
change of kernel width. The computation of $\mathcal{L}$ can be seen as an optimisation problem:
\begin{equation}
	\label{eq_robust_opti}
	\mathcal{L}=\argmax_{l_1, l_2\in(0,+\infty)} \frac{||h_i(l_1)-h_i(l_2)||}{||l_1-l_2||}
\end{equation}
which, unfortunately, is very challenging to solve analytically or to estimate numerically, 
due to the high complexity and non-linearity of the function $h_i$.
Similar difficulties are also 
in, e.g. \citep{alvarezmelis2018robustness}, when studying the robustness to perturbed input instances. 

To bypass the difficulty and still provide insights on the robustness defined earlier, we instead introduce a weaker empirical notion of the robustness to kernel settings:
\begin{definition}
	Assume $L_1$ and $L_2$ are both random variables of kernel width settings within $[l_{lo},l_{up}]$. Then, $R$ is the median\footnote{Although other statistics may also suffice, we choose the median value to cope with the possible extreme outliers.} (denoted as $\mathbb{M}(\cdot)$) of the ratio between the perturbed distances of $h_i$s and the pair $(L_1, L_2)$:
	\begin{equation}
		\label{eq_robust_mean}
		R\!=\!\mathbb{M}\!\left(\!\frac{||h_i(L_1)-h_i(L_2)||}{||L_1-L_2||}  \middle\vert  l_{lo}\!\leq\! L_1\! \leq\! l_{up}, \! l_{lo}\!\leq\! L_2\! \leq\! l_{up}\!\right)
	\end{equation}
which represents the average robustness to the kernel settings when explaining the instance $i$.
\end{definition}
The kernel width cannot be too large nor too small (cf. the general discussions on variance-bias trade-off in selecting hyper-parameters \citep{cawley_over-fitting_2010}). In practice, after a few trials, e.g., by cross-validation, it is not hard to figure out empirically a bound $[l_{lo},l_{up}]$ as the range of all possible kernel settings in which we may perturb the 
kernel width to compute the robustness. In Eq.~\eqref{eq_robust_mean}, we focus on the median value which is a much easier quantity to estimate than the optimised solution in Eq.~\eqref{eq_robust_opti}, yet it provides insights on the general robustness.



\section{Explanation Fidelity}
\label{sec_fidelity}

Among many evaluation criteria on the goodness of explanations, fidelity -- how truthfully the explanation represents the unknown behaviour of the underlying AI model -- is of great importance. In other words, we want the XAI method to explain the \textit{true cause} of the underlying model's prediction \citep{petsiuk2018rise,sun_explaining_2020}. In this paper, we adopt the \textit{actual causality} as an indicator for the explanation fidelity in the following two different ways.

First, we reuse the two causal metrics proposed in \citep{petsiuk2018rise} --  \textit{deletion} and \textit{insertion}. The intuition behind deletion is that the removal of the ``cause'' (important features) will force the underlying AI model to change its decision. To be exact, starting with a complete instance and then gradually removing top-important features (according to the ranked list obtained from XAI methods), the deletion metric measures the decrease in the probability of the predicted label. A sharp drop and thus a low Area Under the probability Curve (AUC), as a function of the fraction of removed features, suggests a good explanation that captures the real causality. Similarly, but in a complementary way, the insertion metric measures the increase in the probability as more and more important features are introduced, with higher AUC indicating 
a better explanation. The two metrics not only alleviate the need for large-scale human annotation effort, but also are better at assessing causal explanations by avoiding human bias \citep{petsiuk2018rise}.

Second, we evaluate explanation fidelity through neural backdoors, 
inspired by 
\citep{lin_evalxai_backdoor_2020,sun_explaining_2020}. 
The major difficulty of evaluating fidelity is the lack of ground truth behaviour of the underlying AI model. 
As the most important features that cause deliberate misclassification, backdoor triggers provide such ground truth and therefore should be highlighted by a good XAI method. 
Both \citep{lin_evalxai_backdoor_2020} and \citep{sun_explaining_2020} use the metric Intersection over Union (IoU) to measure the success in highlighting the trigger. Given a bounding box around the true trigger area $B_T$ and the highlighted area by XAI methods $B'_T$, the IoU is their overlapped area divided by the area of their union, i.e., $(B_T\cap B'_T)/(B_T\cup B'_T)$. IoU is an established metric originally designed for object detection that ranges from 0 to 1, a higher IoU is better. It only considers the overlapping of the highlighted area and the ground truth, ignoring how geometrically closed they are when there is no overlapping. To complement this, we introduce a secondary metric in Appendix \ref{sec_app_pk_from_vnv} to further confirm our conclusions.

\section{A Bayesian Retrofit of LIME}
\label{sec_BayLIME}




Denote the input set with $n$ samples as $\pmb{X}=\{\pmb{x_1},...,\pmb{x_n}\}$ where $\pmb{x_i}$ is the instance $i$ with $m$ features 
(i.e. $\pmb{X}=(x_{ij}) \in \mathbb{R}^{n\times m}$), and let the corresponding $n$ target values be a column vector $\pmb{y}=[y_1,...,y_n]^T$. In Bayesian linear regression, a response $y_i$ is assumed to be Gaussian distributed around $\pmb{x_i}\pmb{\beta}$:
\begin{equation}
	Pr(y_i \mid \pmb{\beta}, \pmb{x_i}, \alpha)=\mathcal{N}(y_i\mid\pmb{x_i}\pmb{\beta},\alpha^{-1})
\end{equation}
where $\pmb{\beta}$ is the coefficient 
vector of $m$ features, and $\alpha$ is the precision parameter (reciprocal of the variance) representing noise in this linear assumption. Then, we may write down the likelihood function: 
\begin{equation}
	\label{eq_likelihood_f}
	Pr(\pmb{y} \mid \pmb{\beta}, \pmb{X}, \alpha)=\prod_{i=1}^n\mathcal{N}(y_i\mid\pmb{x_i}\pmb{\beta},\alpha^{-1}).
\end{equation}
For computational convenience, we choose a conjugate prior distribution for $\pmb{\beta}$ -- another (multivariate) Gaussian:
\begin{equation}
	\label{eq_prior}
	Pr(\pmb{\beta}\mid \pmb{\mu_0}, \pmb{S_0})=\mathcal{N}(\pmb{\beta} \mid \pmb{\mu_0}, \pmb{S_0})
\end{equation}
where $\pmb{\mu_0}$ and $\pmb{S_0}$ are the prior mean vector and covariance matrix, respectively. Then, thanks to the conjugacy,
the posterior $\pmb{\beta}$ is also a Gaussian:
\begin{align}
	Pr(\pmb{\beta}\mid \pmb{y}, \pmb{X},\alpha,\pmb{\mu_0}, \pmb{S_0})
	&\propto Pr(\pmb{y} \mid \pmb{\beta}, \pmb{X}, \alpha)Pr(\pmb{\beta}\mid \pmb{\mu_0}, \pmb{S_0})\nonumber
	\\
	&=\mathcal{N}(\pmb{\beta} \mid \pmb{\mu_n}, \pmb{S_n})
	\label{eq_postrior}
\end{align}
with the  mean vector $\pmb{\mu_n}$ and covariance matrix $\pmb{S_n}$:
\begin{align}
	\label{eq_post_mean_vector}
	\pmb{\mu_n}=\pmb{S_n}(\pmb{S_0}^{-1}\pmb{\mu_0}+\alpha \pmb{X}^T \pmb{y}),\,\,
	\pmb{S_n}^{-1}=\pmb{S_0}^{-1}+\alpha \pmb{X}^T\pmb{X}
\end{align}
For simplicity and without loss of generality, we assume the Gaussian prior is governed by a
single precision parameter $\lambda$, i.e. $\pmb{S_0}=\lambda^{-1}\pmb{I}_{m}$ where $\pmb{I}_m$ is a $m \times m$ identity matrix.


\subsection{Embedding Prior Knowledge in LIME}

The local surrogate model is trained on \textit{weighted} samples perturbed around the instance of interest, with weights representing their respective proximity to the instance. 
Therefore, the posterior estimate on $\pmb{\beta}$ is now as follows. 
\begin{align}
	\pmb{\mu_n}\!=\!\pmb{S_n}(\lambda\pmb{I}_{m}\pmb{\mu_0}+\alpha \pmb{X}^T\pmb{W}\pmb{y}),\,
	\pmb{S_n}^{-1}\!=\!\lambda\pmb{I}_{m}+\alpha \pmb{X}^T\pmb{W}\pmb{X}
	\label{eq_post_cov_matrix_weighted}
\end{align} 
where the weights $\pmb{W}=\mathit{diag(w_1,\dots,w_n)}$ is a diagonal matrix calculated by a kernel function according to the new samples' proximity to the original instance. 
The first equation of \eqref{eq_post_cov_matrix_weighted} can be rewritten as:
\begin{align}
	\label{eq_weighted_sum}
	\pmb{\mu_n}&=
	(\lambda\pmb{I}_{m}+\alpha \pmb{X}^T\pmb{W}\pmb{X})^{-1}
	\lambda\pmb{I}_{m}\pmb{\mu_0}\nonumber
	\\&+(\lambda\pmb{I}_{m} +\alpha \pmb{X}^T\pmb{W}\pmb{X})^{-1}
	\alpha \pmb{X}^T\pmb{W}\pmb{X} \pmb{\beta}_{\mathit{MLE}}
\end{align}
where $\pmb{\beta}_{\mathit{MLE}}=(\pmb{X}^{T}\pmb{W}\pmb{X})^{-1} \pmb{X}^T\pmb{W}\pmb{y}$ is the Maximum Likelihood Estimates (MLE) for the linear regression model \citep{bishop_pattern_2006} on the weighted samples. 

The Eq.~\eqref{eq_weighted_sum} is essentially a \emph{weighted} sum of $\pmb{\mu_0}$ and $\pmb{\beta}_{\mathit{MLE}}$ -- A \textit{Bayesian combination} of prior knowledge and the new observations. The weights are proportional to: (i) $\lambda\pmb{I}_{m}$, the ``pseudo-count'' of prior sample size based on which we form our prior estimates $\pmb{\mu_0}$; (ii) $\alpha\pmb{X}^T\pmb{W}\pmb{X}$, the ``accurate-actual-count'' of observation sample size, i.e. the actual observation of the $n$ new samples $\pmb{X}^T\pmb{W}\pmb{X}$ scaled by the precision $\alpha$.

To see the above insight clearer, we present the special case of a single feature instance ($m=1$) with a simplified kernel function that returns a constant weight $w_c$ (i.e. $w_i=w_c, \forall i=1,\dots,n$), then 
$\mu_n$ ($\pmb{\mu_n}$ with 1 feature) becomes:
\begin{align}
	\label{eq_post_m1}
	\frac{\lambda}{\lambda+\alpha w_c\sum_{i=1}^{n}x^2_i}\mu_0+\frac{\alpha w_c\sum_{i=1}^{n}x^2_i}{\lambda+\alpha w_c\sum_{i=1}^{n}x^2_i}\beta_{\mathit{MLE}}
\end{align}
where $\beta_{\mathit{MLE}}=\frac{\alpha w_c\sum_{i=1}^{n}x_iy_i}{\alpha w_c\sum_{i=1}^{n}x^2_i}=\frac{\sum_{i=1}^{n}x_iy_i}{\sum_{i=1}^{n}x^2_i}$. 
Now, denote $X$ as the r.v. representing $x_i$, we know:
\begin{align}
	\label{eq_approxi_samples}
	\sum_{i=1}^{n}x^2_i\approx n \mathbb{E}(X^2)=n\left(\mathit{Var}(X)+\mathbb{E}(X)^2\right).
\end{align}
As implemented by LIME, $x_i$s are $n$ random samples from some distribution, depending on the type of the feature, e.g.: (i) For superpixels of images, $x_i \in \{0,1\}$ are random samples from an uniform 0-1 two-point distribution; (ii) For numerical features of a tabular dataset, $x_i$ are random samples from a standard Gaussian $\mathcal{N}(0,1)$. Thus, say, for numerical features from tabular data, $\sum_{i=1}^{n}x^2_i \approx n(1+0^2)=n$ via Eq.~\eqref{eq_approxi_samples}. Then, Eq.~\eqref{eq_post_m1} can be simplified as:
\begin{align}
	\label{eq_post_m1_2}
	\frac{\lambda}{\lambda+\alpha w_c n}\mu_0+\frac{\alpha w_c n}{\lambda+\alpha w_c n}\beta_{\mathit{MLE}}
\end{align} which neatly shows how the prior knowledge is embedded in our BayLIME in an intuitive story -- ``based on $\lambda$ data-points prior to our new experiment, we form our prior estimate of $\mu_0$. Now, in the experiments, we collect $n$ samples. After considering the precision ($\alpha$) and weights ($w_c$) of the new samples, we form a MLE estimate $\beta_{\mathit{MLE}}$. Then, we combine the two estimates -- $\mu_0$ and $\beta_{\mathit{MLE}}$ -- according to their proportions of the \textit{effective} samples size used, i.e., $\lambda$ and $\alpha w_c n$, respectively. Finally, the confidence in our new posterior estimate is captured by all effective samples used, i.e. $\lambda+\alpha w_c n$ (the posterior precision)''. Now it is easy to see via Eq.~\eqref{eq_post_m1_2} (also holds for Eq.~\eqref{eq_weighted_sum}):

\begin{remark}
	\label{rem_interp_lambda_alpha}
	Smaller $\lambda$ means lower confidence in the prior knowledge, thus the posteriors are mostly dominated by the new observations, and vice versa. That is, in extreme cases:
\\
1) When $\lambda\simeq 0$ the result \eqref{eq_weighted_sum} (and its simplified case \eqref{eq_post_m1_2}) reduces to MLE, i.e. ``let the data speak for themselves''.
\\
2) On the other hand, if $n\simeq 0$ (or equivalently $\alpha \simeq 0$, $w_c \simeq 0$), then the $\beta_\mathit{MLE}$ estimate from the new data is negligible and the prior knowledge dominates the posteriors.
\end{remark}
This kind of intuitive decomposition is generic to Bayesian inference in the linearly updated canonical-exponential families \citep{bernardo_bayesian_1994}, which is found to be useful in applications that exploit prior knowledge, e.g., \citep{filieri_formal_2012,zhao_probabilistic_2019,walter_imprecision_2009,zhao_interval_2020}.

\begin{remark}
\label{rem_why_prior_helps}
Via the ``weighted sum'' Eq.~\eqref{eq_weighted_sum} (and its simplified case \eqref{eq_post_m1_2}), the general reasons why integrating prior knowledge improve the three aforementioned properties are: 
\\
\textit{1)} $\pmb{\beta}_{\mathit{MLE}}$ is a function of the $n$ randomly perturbed samples that causes inconsistency, while $\pmb{\mu_0}$ is independent from the cause. A weighted sum of both may ``average out'' the randomness, thus improving the consistency.
\\
\textit{2)} $\pmb{\beta}_{\mathit{MLE}}$ is a function of $\pmb{W}$ that depends on the choice of kernel settings, while $\pmb{\mu_0}$ is independent from kernel settings. A weighted sum of both may ``average out'' the effect from kernels, thus improving the robustness.
\\
\textit{3)} $\pmb{\mu_0}$ normally contains added diverse information to $\pmb{\beta}_{\mathit{MLE}}$ (black-box queries) that benefits the explanation fidelity.
\end{remark}

\subsection{The BayLIME Framework}
\label{sec_Baylime_framework}

BayLIME uses Bayesian linear regressors as local surrogate models,
and implements three options as shown in Table \ref{tab_baylime_3_options}.

\begin{table}[h]
\centering
\caption{The BayLIME framework with three options}
\resizebox{1\linewidth}{!}
{
\begin{tabular}{r|c|c|c}
\hline
\textit{BayLIME Opt. \textbackslash \, prior para.}     & $\pmb{\mu_0}$ & $\lambda$ & $\alpha$ \\ \hline
non-informative priors    &    zero vector     &   fitted        &    fitted      \\ \hline
partial informative priors &   known      &    known       &     fitted     \\ \hline
full informative priors   &    known     &     known      &     known     \\ \hline
\end{tabular}
}
\label{tab_baylime_3_options}
\end{table}

\textbf{BayLIME with non-informative
priors.}
When no prior knowledge is available, we assume a zero mean vector for $\pmb{\mu_0}$ and do \textit{Bayesian model selection} \citep[Chpt.~3.4]{bishop_pattern_2006} for $\lambda$ and $\alpha$. Specifically, BayLIME reuses
the established algorithm to fit $\alpha$ and $\lambda$ from data \citep{mackay_bayesian_1992,tipping_sparse_2001}. The solutions for $\lambda$ and $\alpha$ are \textit{implicit}, since they are obtained by starting with initial values and  then iterating over some interval equations until convergence.

\textbf{BayLIME with partial informative priors.} We assume a known complete prior distribution of the feature coefficients with mean vector $\pmb{\mu_0}$ and precision $\lambda$. We call this option in BayLIME ``with partial priors'' in the sense that we still do not know the parameter $\alpha$, for which we fit from data. Similarly as before, we modify the Bayesian model selection algorithm by iterating $\alpha$ (but with fixed $\lambda$ in this case) to maximise the log marginal likelihood until convergence.


\textbf{BayLIME with full informative priors.} By assuming an ideal scenario in which we have full prior knowledge of all the $\pmb{\mu_0}$, $\lambda$ and $\alpha$ parameters, BayLIME may directly implement the closed-form results of 
Eq.~\eqref{eq_post_cov_matrix_weighted}.

\subsection{Ways of obtaining prior knowledge}
\label{sec_obtain_pk}
How to elicit accurate priors for BayLIME varies case by case (depending on the application-specific context), and is indeed challenging.
However, we believe such challenge is 
neither unique to BayLIME, but generic to any approach with a Bayesian flavour, nor 
clueless in practice. That said, we implemented 3 ways of getting priors in our experiments (but not limited to, cf. Sec.~\ref{sec_discussion} for more discussions):


\textit{a)} Validation and Verification (V\&V) methods that directly analyse the behaviour of the underlying AI model may indicate the importance of certain features, yielding priors required by BayLIME. E.g., when explaining a prediction made by an infected model, (imperfect) detection tools may provide prior knowledge on possible backdoor triggers. We implemented this scenario by using the results of the popular NeuralCleanse \citep{wang_neural_2019} as the priors.

\textit{b)} There are emerging XAI techniques (cf. \citep{huang_survey_2020,molnar_interpretable_2020} for a review). Explanations by some other \textit{diverse} XAI explainers based on fundamentally different theories to LIME (e.g., gradient-based vs perturbation-based, global vs local) may provide useful prior knowledge. Presumably, the drawbacks of individual XAI methods can be mitigated by the ``hybrid'' solution provided by BayLIME. We implemented and confirmed this conjecture by using GradCAM \citep{selvaraju2017grad} results as the priors.

\textit{c)} We also use explanations of \emph{a set of similar instances} (to the instance under explanation) to form our prior knowledge. Although it is still not rigorous enough (e.g., how to decide what is a ``similar'' instance), we believe this way of obtaining priors serves as a first illustrative example.
\section{Evaluation}
\label{sec_evaluation}

We evaluate our BayLIME by performing extensive experiments to address the following research questions.


\textbf{RQ1:} How effectively BayLIME improves the consistency of LIME in different scenarios of available prior knowledge?

\textbf{RQ2:} How effectively BayLIME improves the robustness to kernel settings of LIME in different scenarios of available prior knowledge?

\textbf{RQ3:} Will BayLIME improve the explanation fidelity compared to some state-of-the-art XAI methods in different scenarios of available prior knowledge?

\begin{figure*}[h]
	\centering
	\includegraphics[width=1\linewidth]{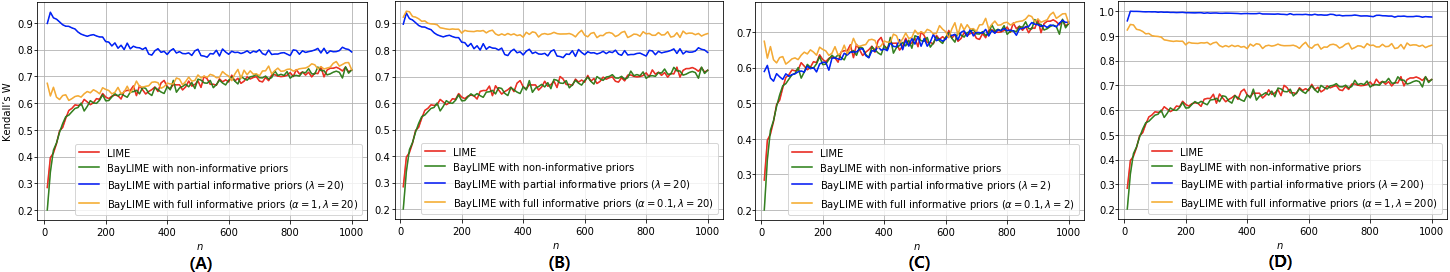}
	\caption{Kendall's~W in $k=200$ repeated explanations by LIME/BayLIME on random Boston house-price instances. Each set shows an illustrative combination of $\alpha$ and $\lambda$. Same patterns are observed on images, cf. Appendix \ref{sec_app_incon_measure} for more.}
	\label{fig_incon_kendallw}
\end{figure*}

\textbf{Evaluation methods.} LIME is one of the few XAI techniques that work for all tabular data, text and images \citep{molnar_interpretable_2020}. Our experiments were also conducted on diverse types of datasets including the Boston house-price dataset, the breast cancer Wisconsin dataset (with both numerical and categorical features), and a set of CNNs pretrained on the ImageNet and GTSRB \citep{stallkamp_man_2012}. While only typical results are presented in the main paper, cf. the appendices and our BayLIME project website\footnote{
\url{https://github.com/x-y-zhao/BayLime}
} for more.

For RQ1, in addition to LIME, we select a set of BayLIME explainers with different options and prior parameters. For each explainer, we iterate the explanation of the given instance $k$ times, and quantify the inconsistency according to Kendall's~W. We treat the sample size $n$ as a variable to confirm our earlier observation that inconsistency is an even severer problem when $n$ has to be limited for efficiency.

To answer RQ2, we firstly define an interval $[l_{lo},l_{up}]$ as the empirical bounds of all possible kernel width settings for the given application. We randomly sample from that interval 5000 pairs of kernel width parameters. Then, for each pair, we can calculate the ``distance'' of the two explanations\footnote{In RQ2, we fix the perturbed sample size to a sufficiently large number, e.g., $n=1000$ (as indicated by the results of RQ1), to minimise the impact of inconsistency.}. Finally, we obtain a sample set of ratios between the two distances of explanations and the kernel width pair, on which statistics provides insights on the general robustness, e.g., the median value as used by Eq.~\eqref{eq_robust_mean}.

The prior knowledge used in RQ1 and RQ2 is obtained from previous LIME explanations on a set of \textit{similar} instances. Specifically, the average importance of each feature in that set collectively forms our prior mean vector $\pmb{\mu_0}$, and the number of similar instances implies $\lambda$, leaving the $\alpha$ either as unknown (then fitted from data) or assigned empirically (based on previous fitting of $\alpha$ on similar instances).

For RQ3, we implemented 2 scenarios in which the prior knowledge is obtained from other diverse XAI and V\&V methods respectively: 

\textbf{RQ3.a)} We first use the gradient-based XAI approach GradCAM to
obtain ``grey-box'' information of the underlying CNN as the prior, and then 
combines 
such extra knowledge with the ``black-box'' evidence as utilised by LIME. We compare the explanation fidelity of BayLIME with LIME, GradCAM and SHAP, via the deletion and insertion metrics. 

\textbf{RQ3.b)} In this scenario, we aim to explain the behaviour of infected CNNs (BadNet \citep{gu_badnets_2019} and TrojanAttack \citep{liu_trojaning_2018} both trained on GTSRB with backdoors). The backdoor detector NeuralCleanse is used to \textit{approximate} the location of triggers first. Then, BayLIME considers such knowledge as priors to provide a better explanation on the attacking behaviour, compared to NeuralCleanse and LIME applied solely. For the two example scenarios, we choose the relatively more practical option -- BayLIME with partial priors, 
i.e., $\alpha$ is fitted, and the readers are referred to 
Appendix \ref{sec_appen} for the formulae with intuitive rationales on eliciting $\pmb{\mu_0}$ and $\lambda$.


\textbf{Results and Discussion.}
\textbf{RQ1.}
The red curves in Fig.~\ref{fig_incon_kendallw} present the Kendall's W measurements as a function of the perturbed sample size $n$. Although it increases quickly, we observe very low consistency when $n$ is relatively small (e.g., $n<200$). 
These results support our earlier conjecture on the inconsistency issue of LIME, especially when $n$ has to be limited by an upper-bound due to efficiency considerations. Non-informative BayLIME is indistinguishable from LIME, since both of them are \textit{only} exploiting the information from the $n$ samples that generated randomly -- the presence of randomness means that the results of sampling cannot be duplicated if the process were repeated. Naturally, the more sparse the samples are, the greater randomness presents in the dataset. Thus, LIME and non-informative BayLIME show \textit{monotonic trends} as $n$ increases.

By contrast, BayLIME with partial/full informative priors can ``average out'' the sampling noise in the new generated data by combining the information from the priors, in a Bayesian principled way of Eq.~\eqref{eq_weighted_sum}. To closely inspect how effectively BayLIME with different priors affects the consistency, we first need the auxiliary of the factor $\lambda/\alpha$. It is a known result that $\lambda/\alpha$ can be treated as a \textit{regularization coefficient} in Bayesian linear regressors \citep[Chpt.~3.3]{bishop_pattern_2006}, meaning a larger $\lambda/\alpha$ penalises more on the training data (to control over-fitting). Indeed, this aligns well with our Remark~\ref{rem_interp_lambda_alpha}: (i) when $\alpha \simeq 0$, the factor $\lambda/\alpha \rightarrow +\infty$ meaning a huge penalty on the data, thus the prior knowledge effectively dominates the posteriors; (ii) when $\lambda \simeq 0$, $\lambda/\alpha \rightarrow 0$ meaning no penalty on the data, thus the posteriors is dominated by the new observations.

Both the plots of BayLIME with full informative priors (yellow curves) in Fig.~\ref{fig_incon_kendallw} (A) and (C) have a regularization factor $\lambda/\alpha=20$, and are basically identical. This is because, once $\lambda/\alpha=20$ is fixed, the ``proportion'' of contributions to the posteriors by the priors and the new data is fixed. In other words, given $n$ samples, the ability of ``averaging out'' sampling noise by the priors is fixed. When $\lambda/\alpha$ increases to 200, as shown by the yellow curves in Fig.~\ref{fig_incon_kendallw} (B) and (D), such ability of averaging out sampling noise is even stronger, which explains why Kendall's W measurements in this case are higher than the case of $\lambda/\alpha=20$. 
For BayLIME with partial informative priors (blue curves in Fig.~\ref{fig_incon_kendallw}), we observe that smaller $\lambda$ results in worse consistency, e.g., Fig.~\ref{fig_incon_kendallw} (C) vs (D). Again, Remark \ref{rem_interp_lambda_alpha} applies here -- smaller $\lambda$ implies less contributions from the priors to the posteriors, meaning with less ability to average out the randomness in the samples.

Starting from a non-zero small number, as $n$ increases, we can see BayLIME with partial/full informative priors may exhibit an uni-modal pattern, e.g., the Fig.~\ref{fig_incon_kendallw} (C)\footnote{Other plots may not show the pattern clearly due to the range/scale of the axes and inevitable noise in the experiments.} with a minimum point. This represents a tension between the \textit{perfectly consistent prior knowledge} (does not change at all in repeated explanations) and \textit{quite consistent MLE based on large samples}. There must be a ``balance-point'' in-between that compromises both ends of the tension, yielding a minimised consistency. Finally, when $n \rightarrow +\infty$, it is trivial to see (e.g., by taking the limit of Eq.~\eqref{eq_post_m1_2} as a function of $n$), all plots will eventually converge (to the measurement based on MLE using infinitely large samples).



\begin{figure}[h]
	\centering
	\includegraphics[width=1\linewidth]{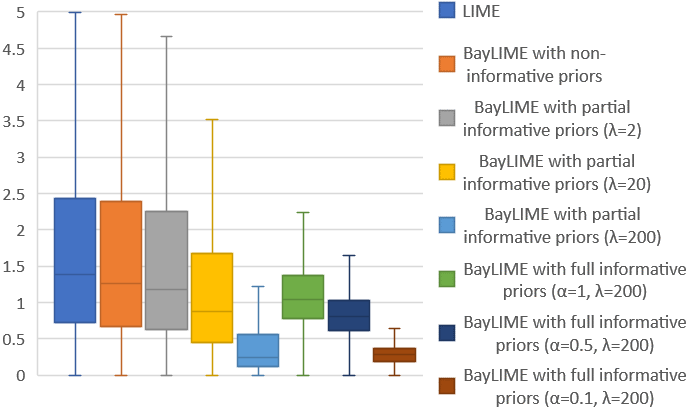}
	\caption{The general (un-)robustness of eight AI explainers to kernel settings (box-and-whisker plots without outliers).}
	\label{fig_box_plot_robust}
\end{figure}

\textbf{RQ2.} Fig.~\ref{fig_box_plot_robust} are box-and-whisker charts providing insights on the general robustness of eight AI explainers to kernel width settings, in which the median values defined by Eq.~\eqref{eq_robust_mean} are marked, as usual, by horizontal bars inside the boxes.

\begin{figure}[h]
	\centering
	\includegraphics[width=1\linewidth]{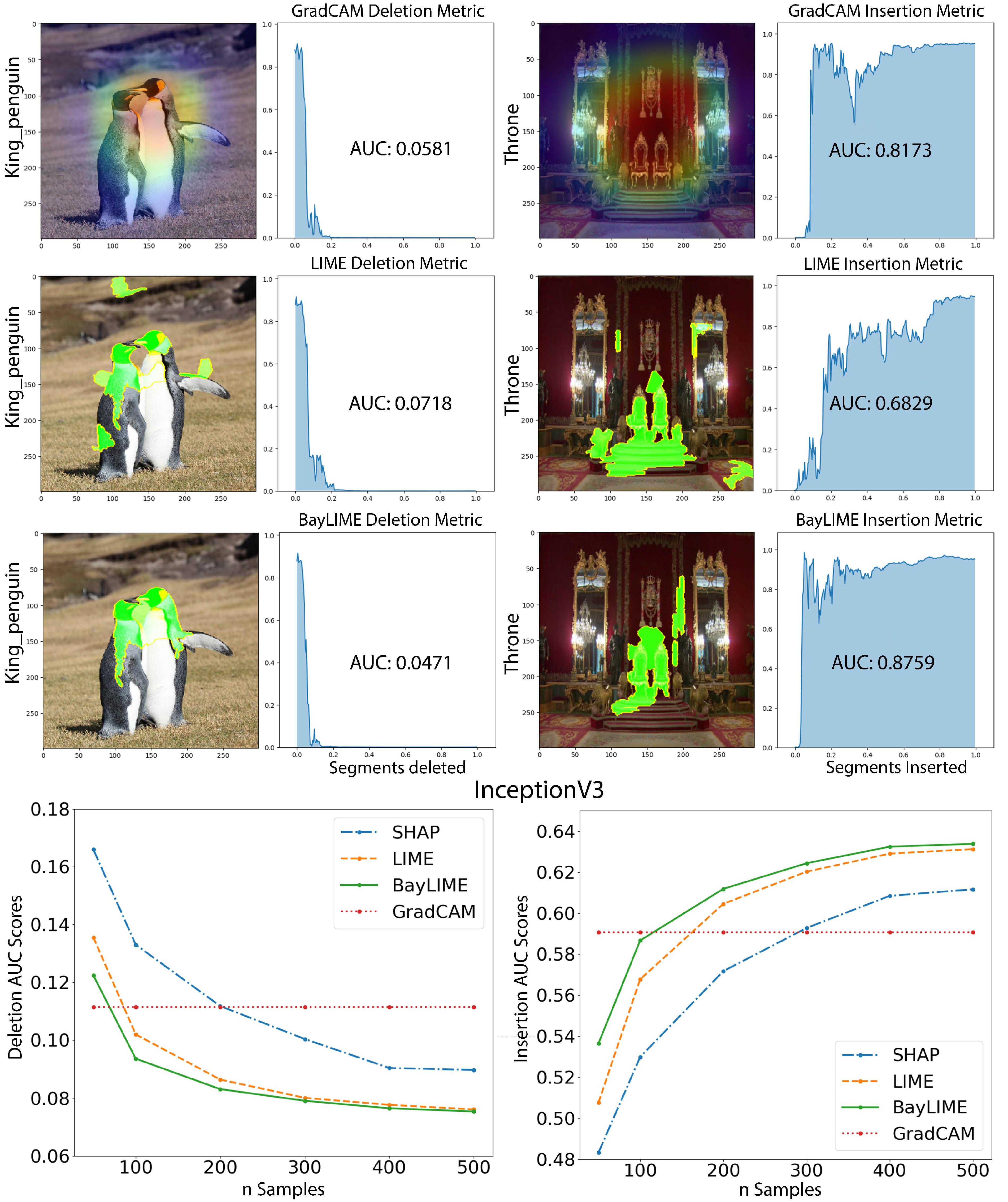}
	\caption{Two sets of examples ($n=300$), and the average AUC (based on 1000 images per value of $n$) of the deletion (smaller is better) and insertion (bigger is better) metrics of BayLIME comparing with other XAI methods.}
	\label{fig_delet_inser_exp}
\end{figure}

Again, LIME and BayLIME with non-informative priors exhibit similar robustness, since there is no prior knowledge being considered, rather the data solely determines the explanations of both. In stark contrast, when either partial or full prior knowledge is taken into account, we observe an obvious improvement on the robustness to kernel settings.

The regularisation factor $\lambda/\alpha$ and Remark~\ref{rem_interp_lambda_alpha} are still handy here in the discussions on how varying the $\lambda$ and $\alpha$ affects the robustness -- it all boils down to how much contribution from the priors (that is independent from kernel setting), compared with the contribution from the new data (that is sensitive to kernel settings), to the posteriors.


\begin{figure}[h!]
	\centering
	\includegraphics[width=1\linewidth]{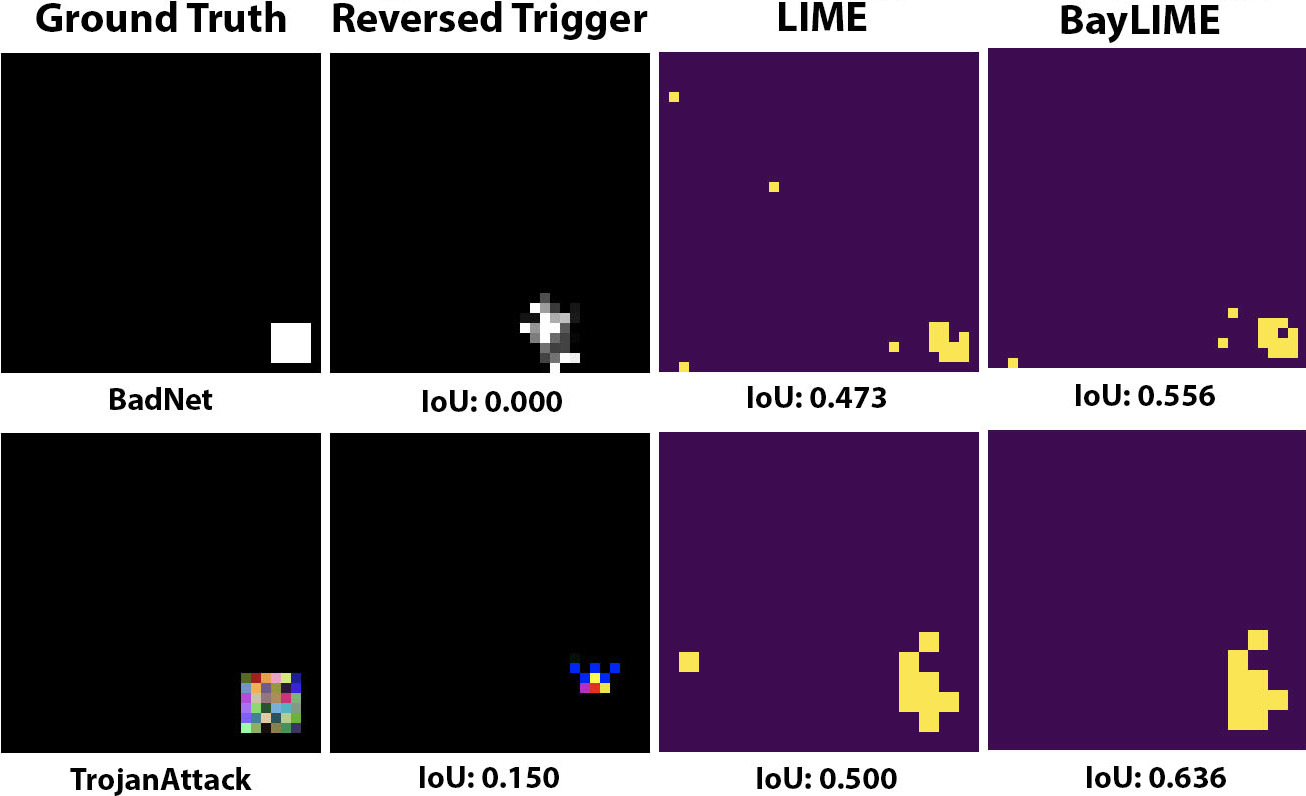}
	\caption{Two examples and their IoU measurements of BayLIME using V\&V results as priors (reversed triggers by NeuralCleanse), compared to the results of NeuralCleanse and LIME applied individually. LIME and BayLIME use the top-few features  (where each feature is a fixed size square superpixel, cf. Appendix \ref{sec_app_pk_from_vnv} for details) such that the total number of pixels is the same as the ground truth. NB: We omit the background image to highlight the triggers only.}
	\label{fig_vnv_example}
\end{figure}

\textbf{RQ3.a.}
GradCAM 
provides fundamentally different 
grey-box information as priors to the balck-box LIME. BayLIME, as expected,  performs better than both GradCAM (the prior) and LIME (new observations), thanks to its unique advantage of incorporating diverse information. Fig.~\ref{fig_delet_inser_exp} first shows two sets of such examples via the AUC scores of the deletion and insertion metrics, respectively. Then statistics on the average scores (varying $n$) are shown in the last row of Fig.~\ref{fig_delet_inser_exp}. We observe: (i) BayLIME performs better than SHAP and LIME, while there is a converging trend when $n$ increases. This aligns with the second point in Remark~\ref{rem_interp_lambda_alpha}. (ii) GradCAM is not a function of $n$ (thus showing horizontal lines) and only performs better in the corner cases when $n$ is extremely small (even smaller than the number of features). 
We conclude that, compared to the other 3 XAI methods, BayLIME has better fidelity in the middle and most practical range of $n$ (e.g., 100$\sim$400 in our case). For more examples/statistics on other CNNs cf. Appendix \ref{sec_app_pk_from_gradcam}


\textbf{RQ3.b.} The V\&V tool NeuralCleanse yields reversed triggers as the approximation of backdoor, which are far from perfect. E.g., in the BadNet example of Fig.~\ref{fig_vnv_example} (1st row), the reversed trigger completely missed the ground truth trigger (thus $\text{IoU}=0$), despite being very closed to. Even directly apply LIME on an attacked image may provide a better IoU than NeuralCleanse. Nevertheless, BayLIME performs the best
after considering both the reversed triggers and a surrogate model as LIME (trained on the same number of $n$ samples). As shown in Table \ref{tab_iou_statistics}, statistics on averaging the IoU of 500 random images confirm the above observation (NeuralCleanse is independent from individual images).

\begin{table}[h]
	\centering
	\caption{Statistics on the average IoU by 3 methods.} 
	\resizebox{1\linewidth}{!}
	{
		\begin{tabular}{c|ccc}
        \hline
        Model
        & NeuralCleanse & LIME & BayLIME \\ \hline
        BadNet & 0.000 & 0.385 & \textbf{0.406}  \\
        TrojanAttack & 0.150 & 0.599 & \textbf{0.637} \\ \hline
\end{tabular}}
\label{tab_iou_statistics}
\end{table}

\section{Related Work}
\label{sec_related_work}

Several prior works aim at improving aspects of LIME: 
\citep{shi_modified_2020} modify the perturbed sampling method of LIME to cope with the correlation between features. RL-LIM \citep{Yoon_rllime_2019} employs reinforcement learning to select a subset of perturbed samples to train the local surrogate model. MAPLE \citep{plumb_model_2018} uses local linear regressors along with random forests that determines weights of perturbed instances to optimise the local linear model. Both DLIME \citep{zafer_dlime_2019} and ALIME \citep{shankara_alime_2019} share the similar concern with us on the inconsistency of repeated explanations. DLIME replaces the random perturbation with deterministic methods and ALIME employs an auto-encoder as a better weighting function for the local surrogate model. In comparison to these, BayLIME seeks inherently different solutions in a Bayesian way which has the unique advantage of embedding prior knowledge (thus direct experimental comparisons between them are not sensible -- prior knowledge plays an important role in BayLIME that cannot be represented by other two methods). Moreover, BayLIME can deal with strict efficiency constraints (i.e., with small perturbed sample size $n$) and may improve explanation fidelity as well, while DLIME and ALIME cannot.


To the best of our knowledge, the only two model-agnostic XAI techniques with a Bayesian flavour
are in \citep{guo_explaining_2018,slack_how_2020}. The former derives generalised insights for an AI model through a \emph{global} approximation (by a Bayesian non-parametric regression mixture model with multiple elastic nets). It fits conjugate priors from data, and normally requires a large sample size for the inference on a large number of model parameters. \cite{slack_how_2020}, in an approach developed independently from BayLIME, uses the posterior credible intervals to determine an ideal sample size $n$. In contrast to both, BayLIME is the first to \textit{exploit prior knowledge} for better consistency, robustness to kernel settings and explanation fidelity.


\section{Discussion}
\label{sec_discussion}

Prior knowledge plays a key role in BayLIME, thus we discuss the following questions on priors to highlight its practical usefulness.

\textbf{Where can we obtain the priors?} 
The answer to this question varies case by case, depending on the application-specific context. We have implemented three typical ways of obtaining priors, i.e. from other diverse XAI tools, V\&V methods, and previous explanations of similar instances. More practical ways of getting priors for BayLIME are certainly
possible and we plan to investigate more in future.


\textbf{How do we know if the prior is good?}
First of all, we need to 
be clear about the definition of ``good'' priors -- is it the prior that truly/accurately reflects the unknown behaviour of the AI model? Or the prior gives an explanation that looks good to human users? The latter definition is ``cheating''  in a Bayesian sense -- we cannot rig a prior to get a result we like -- and this is not the purpose of XAI methods. The former is sensible, but due to the lack of {ground truth} behaviours of the black-box AI model, we would never know the prior is good or not \emph{with certainty}. However, we can be \emph{fairly confident} that a given prior is good in some practical cases (e.g., derived from a V\&V tool that was shown to be reliable in previous uses), so that it can be utilised by BayLIME.

\textbf{What if we used a bad prior?} Reusing the sensible definition of good priors above, indeed, there could be the case we are using a bad prior that introduces bias. In this case, BayLIME might end up providing an explanation that is ``consistently and robustly bad'', which seems to imply that BayLIME is only useful when we are {certain} that the priors are good. However, in practice, we can never be certain if a prior is good or bad, rather \emph{the prior is simply a piece of evidence} to us. It is not only against the Bayesian spirit but also unwise to discard any evidence {without sound reasons}.  If there \textit{is} a proof that the evidence (either the priors or the new observations) is not trustworthy, then certainly we should not consider it in our inference -- in this sense, all Bayesian methods depend on good priors, not just BayLIME. So with some caution in deriving priors/evidence from trustworthy ways (a separate engineering problem that is normally out of the scope of a Bayesian method itself), we may ease the concern of BayLIME being ``consistently and robustly bad''.



\section{Conclusions and Future Work}
\label{sec_conclusion}

This paper introduces BayLIME, a principled, Bayesian based method to integrate useful prior knowledge into the well-known XAI method LIME. Our theoretical analysis and extensive experiments show that the Bayesian mechanism helps in improving the consistency in repeated explanations of a single prediction, robustness to kernel settings, and the explanation fidelity.

Embedding diverse sources of knowledge has been a clear trend in AI in recent years, as the effort to make black-box learning methods more transparent and interpretable. 
Emerging techniques peeking inside AI models 
will lead to many
different ways of obtaining knowledge of their black-box behaviour. 
Therefore, principled methods to integrate diverse knowledge are called for. In future work, we will investigate more practical ways of obtaining priors and how to leverage the posterior confidence for better explanations. Last but not least, we will develop novel use cases of BayLIME.

\begin{acknowledgements} 
This project has received funding from the European Union’s Horizon 2020 research and innovation programme under grant agreement No 956123. 
This work is supported by the UK EPSRC (through the Offshore Robotics for Certification of Assets [EP/R026173/1] and End-to-End Conceptual Guarding of Neural Architectures [EP/T026995/1]) and the UK Dstl (through the project of Safety Argument for Learning-enabled Autonomous Underwater Vehicles). XZ’s contribution is partially supported through Fellowships at the Assuring Autonomy International Programme.
\end{acknowledgements}


\bibliography{ref}

\clearpage
\appendix
\section{Appendix: Supplementary Materials for RQ3}
\label{sec_appen}

In this section, we present the design details and more results of the experiments for \textbf{RQ3}. All experiments are conducted on a Macbook Pro 64 bit machine with Intel 2.3GHz 8 cores i9 CPU and 16GB RAM.

\subsection{RQ3.a: Prior knowledge from other XAI methods}
\label{sec_app_pk_from_gradcam}

In this example scenario, we use 
the ``partial informative priors''
option of BayLIME.
The prior parameters $\pmb{\mu_0}$ and $\lambda$ are elicited from the results of GradCAM, while the $\alpha$ is fitted from data.

\paragraph{To determine $\pmb{\mu_0}$,} we take the following steps:

\textit{a)} Obtain the heatmap (generated by GradCAM) in which each pixel of the image instance has a gradient-value as its importance in the explanation.

\textit{b)} Do segmentation on the image (using the LIME's default algorithm), then calculate the importance of each superpixel by averaging the importance of all pixels inside the superpixel. This step essentially bridges the gap between pixel-wised heatmap generated by GradCAM and superpixel-wised importance vector used by BayLIME.

\textit{c)} Map the gradient-value based importance vector (derived from GradCAM) and the coefficient-value based importance vector (derived from linear surrogate models) into a unified scale. While there are alternative ways of doing the scaling, we apply the \textit{maximum absolute value scaling} method based on the same scale of the coefficient vector.

\textit{d)} Finally, the \textit{superpixel-wised} and \textit{scaled} importance vector (originally derived from GradCAM) is used as $\pmb{\mu_o}$.

\paragraph{To determine $\lambda$,} we first present some model-structure information of the three CNNs used in our experiments. As summarised in Table \ref{table:settings}, for each CNN, we are showing the dimensions of the input layer and the last convolutional layer, based on which the $\lambda$ can be calculated via Eq.~\eqref{eq_app_lambda_gradcam}. To be exact, 
suppose the input image dimension is $i \times j$, the activation map of last convolutional layer is $l \times w \times d$, where $l$, $w$, and $d$ are the length, width and depth dimension. $\lambda$ is then calculated by
\begin{equation}
\label{eq_app_lambda_gradcam}
    \lambda = \frac{\sqrt{l\cdot w}}{\sqrt{i \cdot j}} d
\end{equation}
Intuitively Eq.~\eqref{eq_app_lambda_gradcam} says:

\textit{a)} GradCAM calculates the heatmap based on a number of $d$ feature maps, and each feature map is of size $l \times w$.

\textit{b)} Each feature map represents an abstraction of some features on the original $i\times j$ image. Therefore, it is ``equivalent'' to a single perturbed sample (in LIME) with some features switched on and off.

\textit{c)} Based on this $d$ number of feature maps (or ``equivalent perturbed samples''), we form our prior estimates.

\textit{d)} However, perturbed samples in LIME are weighted according to the distance to the original image. To mimic such weights of samples, we also associate all feature maps with a factor of $\frac{\sqrt{l\cdot w}}{\sqrt{i \cdot j}}$. 

\textit{e)} Finally, the prior precision parameter $\lambda$, as the \textit{weighed pseudo-count of prior samples} based on which we form the prior estimates (cf. Sec.~\ref{sec_Baylime_framework}), can be elicited via Eq.~\eqref{eq_app_lambda_gradcam}.

\begin{table}[ht]
	\centering
	\caption{Three CNNs under explanation and the configurations of $\lambda$ for BayLIME with partial informative priors.} 
	\resizebox{\linewidth}{!}
	{
		\begin{tabular}{c|c|c|c}
			\hline
			Models & InceptionV3 & Xception & ResNet-50 \\ \hline
			Input Layer & $299\times299$ & $299\times299$ & $224\times224$\\
			Last Convolutional Layer & $8\times8\times2048$ & $10\times10\times2048$ & $7\times7\times2048$ \\ \hline
			Partial Informative Priors & $\lambda = 54.8$ & $\lambda = 68.5$ & $\lambda = 64$\\ \hline
		\end{tabular}
		\label{table:settings}
	}
\end{table}



\paragraph{Experimental results on the fidelity of BayLIME using GradCAM explanations as the priors.} As described in Sec.~\ref{sec_fidelity}, we introduce the \textit{deletion} and \textit{insertion} metrics to study explanation fidelity. We compare BayLIME with three other state-of-the-art XAI methods -- LIME, GradCAM and the kernel SHAP -- on their fidelity in explaining the behaviours of three popular CNNs (cf. Table \ref{table:settings}) trained on the dataset ImageNet. We vary the perturb sample size $n$ in the experiments (i.e. $n$ takes $50,100,200,300,400,500$). 

Similar as LIME, BayLIME allows the selection of top few features by the K-LASSO algorithm, while for brevity, we select all $m$ features when $n>m$ and top-5 features when $n \le m$.

\begin{figure*}[h]
	\centering
	\includegraphics[width=1\linewidth]{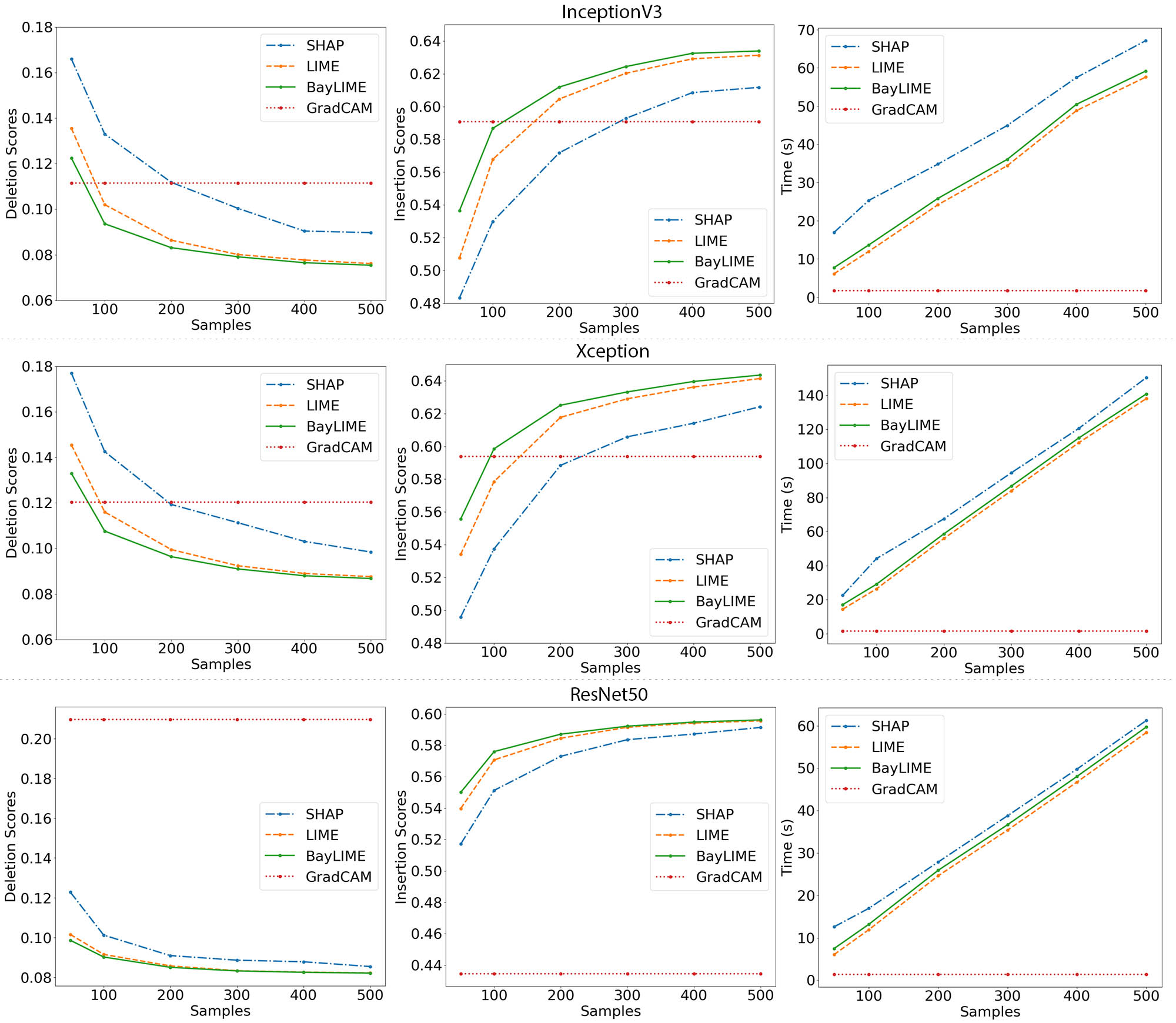}
	\caption{``Deletion and Insertion'' evaluation of LIME, GradCAM, kernel SHAP and BayLIME, based on 1000 randomly selected images from ImageNet per sample size value $n$. The three CNNs InceptionV3, Xception and ResNet50 are studied.}
	\label{fig_ins_del_summary}
\end{figure*}

\begin{figure*}[h]
	\centering
	\includegraphics[width=1\linewidth]{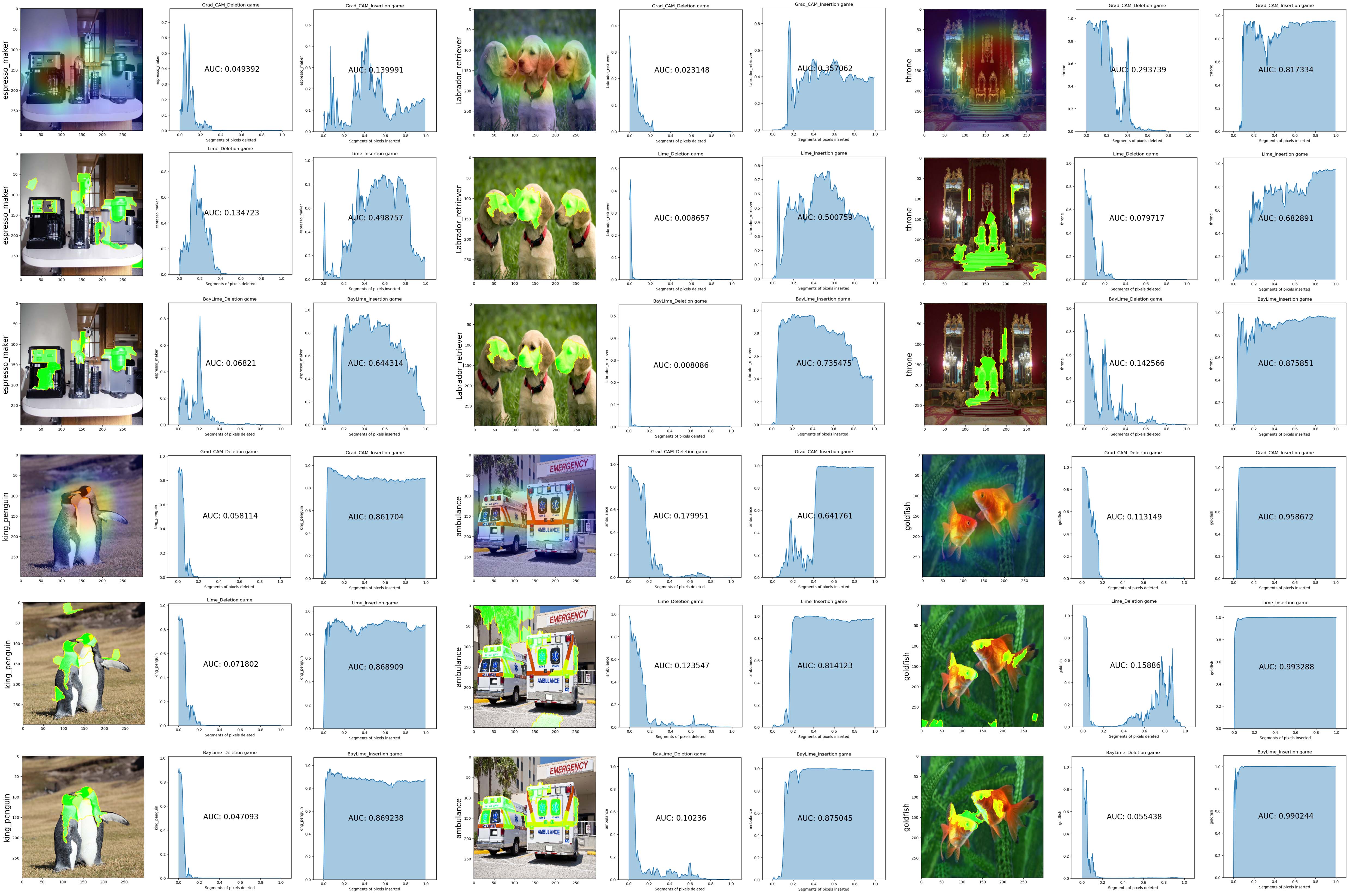}
	\caption{Six sets of examples on the explanations with deletion and insertion measurements (AUC scores). Rows of each set are results from GradCAM, LIME and BayLIME respectively.}
	\label{fig_more_examples}
\end{figure*}

In addition to the Fig.~\ref{fig_delet_inser_exp} presented in the main paper, Fig.s~ \ref{fig_ins_del_summary} and \ref{fig_more_examples} show more statistics and concrete examples. They essentially confirm our conclusions made in Sec.~\ref{sec_evaluation}. Nevertheless, it is worth noting the following few additional observations:

\textit{a)} GradCAM has several inherent problems \citep{chattopadhay_grad_cam_plus_2018}, thus it generally performs worse compared to others with larger sample size (e.g., $n>100$ in our cases). Especially for ResNet50, our experiments show that GradCAM performs noticeably 
worse, 
even when comparing to other explainers in the case when they are working
with smaller sample size. We conjecture that this is due to the known major drawback of GradCAM -- the up-sampling process to create the coarse heatmap results in artefacts and loss of signal. ResNet50 has a severer problem here since it requires a relatively more precise up-sampling process from $7\times 7 \rightarrow 224\times 224$, compared to InceptionV3 and Xception.

\textit{b)} The time cost of BayLIME is effectively a sum of running GradCAM (to get the priors) and LIME (to train local surrogate models). GradCAM is quite light-weight that takes little and constant computational time. Benefiting from this, BayLIME takes slightly more time than LIME that is negligible. Thus, we may claim BayLIME is not less computational efficient than LIME, while improves the fidelity.

\textit{c)} Kernel SHAP (with default hyper-parameters) performs noticeably worse than LIME and BayLIME in all our experiments in terms of both fidelity metrics and computational time. Thus, we omit its examples in Fig.~\ref{fig_more_examples}.

\subsection{RQ3.b: Prior knowledge from V\&V methods}
\label{sec_app_pk_from_vnv}

In this example scenario, we still use BayLIME with partial informative priors. The prior parameters $\pmb{\mu_0}$ and $\lambda$ are elicited from the results of NeuralCleanse, while the $\alpha$ is fitted from data.

\begin{figure}[h!]
	\centering
	\includegraphics[width=1\linewidth]{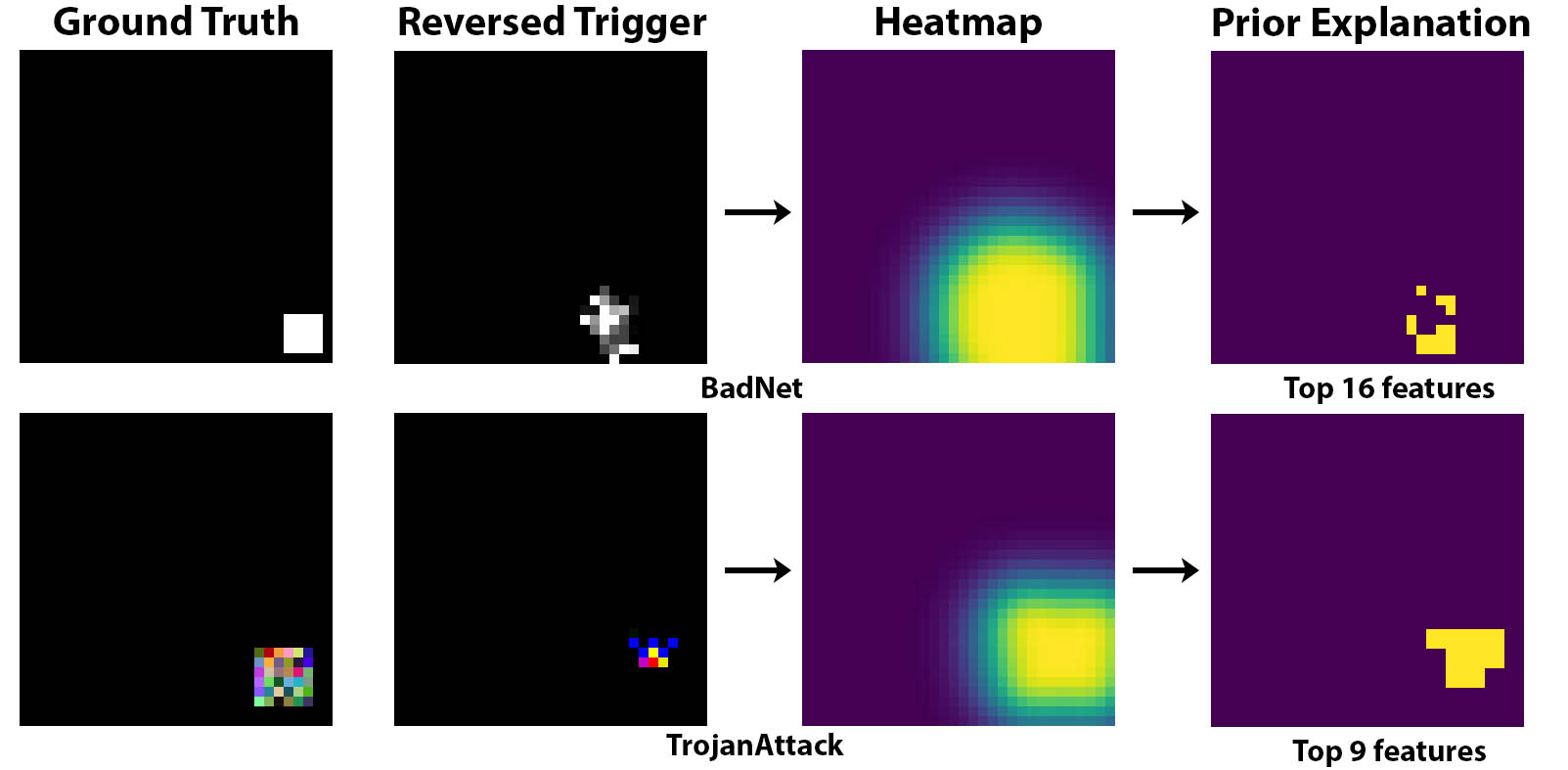}
	\caption{Extract the prior knowledge from reversed triggers.}
	\label{fig_extract_prior}
\end{figure}

\paragraph{To determine $\pmb{\mu_0}$,} we have the following steps:

\textit{a)} Apply NeuralCleanse to generate the reversed triggers, as shown by the second column in Fig.~\ref{fig_extract_prior} which are the reversed triggers of BadNet \citep{gu_badnets_2019} and TrojanAttack \citep{liu_trojaning_2018} respectively.

\textit{b)} Convert the reversed triggers into
heatmaps. Reversed triggers are approximations on the location of the ground truth backdoor. NeuralCleanse deliberately identifies a minimised trigger as such approximation. Consequently, if we directly use the reversed trigger as our priors, then the importance of pixels that close to the trigger would be ignored. Thus, we use radial basis function to generate the heatmap by interpolating the importance values around trigger patterns, cf. the third column in Fig.~\ref{fig_extract_prior}. 

\textit{c)} Do segmentation on the image. Instead of using the LIME's default algorithm here, we deliberately apply $1\times 1$ and $2\times 2$ square superpixels (for the two infected CNNs respectively) that are more common in backdoor attacking scenarios.

\textit{d)} Upon obtaining the heatmap and segmentation of superpixels, the rest of the steps are essentially the same as using GradCAM for the priors (cf. Appendix \ref{sec_app_pk_from_gradcam}).


\paragraph{The determination of $\lambda$.} $\lambda$ is calculated based on the anomaly detection intermediate step in the NeuralCleanse method. Basically, NeuralCleanse first obtains the reversed trigger for each class of labels. Next, each reversed trigger is associated with a $L_1$ norm value. Then the set of all $L_1$ norm values forms a distribution, on which an anomaly detection process is applied to identify the smallest outlier on that distribution, i.e. the minimal trigger is treated as the approximated backdoor. Thus, we set $\lambda$ as the cardinality of the set of $L_1$ norm values that used for the anomaly detection (which is also equal to the total number of classes).

\paragraph{Experimental results on the fidelity of BayLIME using reversed triggers (generated by NeuralCleanse) as the priors.} Since we know the ground truth in this scenario, we adopt the Intersection over Union (IoU) metric to study the explanation fidelity (cf. Sec.~\ref{sec_fidelity}). The 2 infected CNNs are trained on the GTSRB dataset \citep{stallkamp_man_2012}.

In Table \ref{table_iou_and_amd}, we present the statistics of IoU based on 500 backdoor attacked images. Apparently, BayLIME gets the best IoU scores for both infected models, confirming our conclusions in the main paper.

However, as discussed by the end of Sec.~\ref{sec_fidelity}, IoU only considers the overlapping of the highlighted area and the ground truth, ignoring how geometrically closed they are when there is no overlapping. We believe the distance between the highlighted area and the ground truth may also provide some insights on the fidelity, thus we develop a complementary metric to the IoU, called Average Minimum Distance (AMD). AMD measures how close the explanation is to the ground truth:
\begin{definition}
Suppose we get top $k$ important features (or superpixels) from the explainer for image with size $m \times m$. Then, for each superpixel, we can calculate the geometric centre point $(x_i, y_i), i= 1, ..., k$. The AMD from the ground truth pattern $\mathcal{G} = \{(\Tilde{x}_i,\Tilde{y}_i) | i = 1, \dots, \Tilde{k} \}$ is to calculate:
\begin{equation}
    AMD = \frac{\sum_{i = 1}^{k} \min \left(\left(x_i,y_i\right), \mathcal{G}\right) } {k \cdot \sqrt{2(m-1)^2}}
\end{equation}
\end{definition}
We scale the distance between the explanation and ground truth by dividing the maximum distance $\sqrt{2(m-1)^2}$, so that AMD is normalised to $[0,1]$, irrespective of image dimensions.

Now as shown in the AMD column of Table \ref{table_iou_and_amd}, BayLIME again shows better performance than LIME, although the prior by NeuralCleanse shows the best performance. This is unsurprising due to the unique character of NeuralCleanse that only a ``minimised trigger'' will be identified to be as closed to the ground truth as possible, but at the risk of completely missing it (i.e., without any overlapping as reflected by IoU).

\begin{table}[ht]
	\centering
	\caption{Statistics on IoU (higher is better) and AMD (smaller is better) based on 500 backdoor-attacked images. The priors (derived from reversed triggers) are shown in Fig.~\ref{fig_extract_prior}.} 
	\resizebox{\linewidth}{!}
	{
		\begin{tabular}{c|ccc|ccc}
        \hline
        \multirow{2}{*}{Model} & \multicolumn{3}{c|}{IoU} & \multicolumn{3}{c}{AMD} \\
        & Prior & LIME & BayLIME & Prior & LIME & BayLIME \\ \hline
        BadNet & 0.000 & 0.385 & 0.406 & 0.121 & 0.212 & 0.178 \\
        TrojanAttack & 0.150 & 0.599 & 0.637 & 0.011 & 0.072 & 0.049 \\ \hline
\end{tabular}
\label{table_iou_and_amd}
    }
\end{table}

\section{An inconsistency metric with more experiments to RQ1}
\label{sec_app_incon_measure}

\begin{figure*}[h!]
	\centering
	\includegraphics[width=1\linewidth]{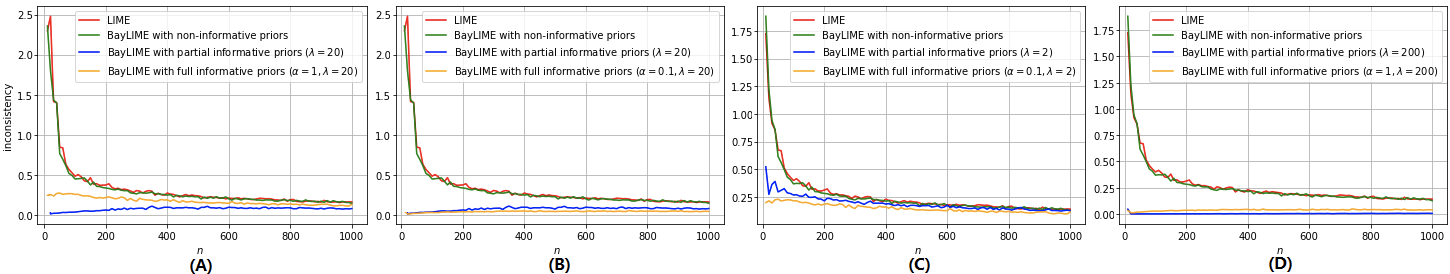}
	\caption{To complement Kendall's~W, the inconsistency metric Eq.~\eqref{eq_incon_def} in $k=200$ repeated explanations by LIME and BayLIME on tabular data. Each set shows an illustrative combination of $\alpha$ and $\lambda$.}
	\label{fig_incon}
\end{figure*}
\begin{figure}[h!]
\centering
\includegraphics[width=1\linewidth]{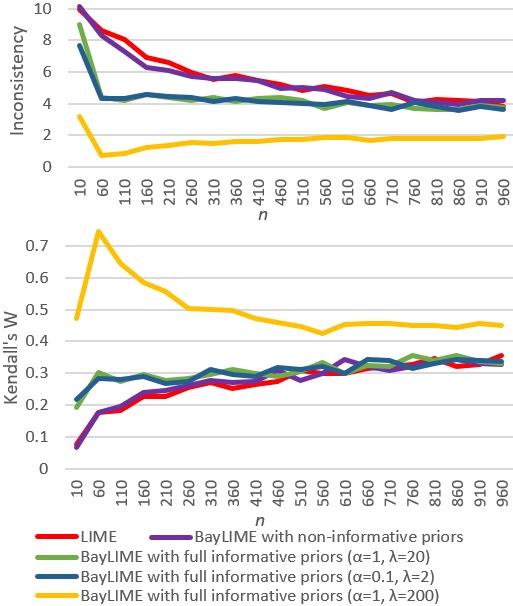}
\caption{For different explainers, the inconsistency of \eqref{eq_incon_def} and Kendall's W in repeated explanations of images labelled by InceptionV3 as functions of the perturbed sample size $n$.}
\label{fig_incon_kenw_image}
\end{figure}

While a quick inconsistency example is shown in Fig.~\ref{fig_lime_time_inconsist} (C)-(E), we introduce a new metric to measure such inconsistency in the following steps :

\textit{a)} In a single explanation, we first normalise the coefficient vector of $m$ features  (represented by $\pmb{\beta}$, cf. the step 6 in the LIME workflow in Appendix \ref{sec_app_preliminaries}) into a unit vector $\pmb{\hat{\beta}}=\frac{\pmb{\beta}}{||\pmb{\beta}||}$. 

\textit{b)} For repeated $k$ explanations (denoted by the upper index $^{(k)}$ in the following notations), the normalised importance of a feature $i$ (i.e. $|\hat{\beta_i}|$) and its ranking may form two probability distributions, denoted respectively as $g_i^{(k)}: [0,1] \rightarrow [0,1]$ and  $f_i^{(k)}:[1,m]\rightarrow [0,1]$. 

\textit{c)} The \textit{index of dispersion} (IoD) of $f_i^{(k)}$ essentially measures the relative variability of the ranks of the feature $i$ in the $k$ repeated explanations, denoted as $\mathit{IoD}(f_i^{(k)})$. To be consistent, we would expect a small $\mathit{IoD}(f_i^{(k)})$. 

\textit{d)} The average importance of each feature $\mathbb{E}(g_i^{(k)})$ is quite different.
Intuitively, we need to weight each $\mathit{IoD}(f_i^{(k)})$ accordingly.

\textit{e)} Finally, we define:
\begin{definition}
	The inconsistency of repeated $k$ explanations for an instance with $m$ features is measured by:
	\begin{equation}
		\label{eq_incon_def}
		\sum_{i=1}^{m} \frac{\mathbb{E}(g_i^{(k)})}{\sum_{j=1}^m \mathbb{E}(g_j^{(k)})} \mathit{IoD}(f_i^{(k)})
	\end{equation}
	where $g_i^{(k)}: [0,1] \rightarrow [0,1]$ and  $f_i^{(k)}:[1,m]\rightarrow [0,1]$ are probability distributions of the normalised importance of feature $i$ and its ranks in $k$ explanations, respectively.
\end{definition}

Intuitively, the inconsistency is a weighted sum of IoD of each feature's ranks in repeated explanations, so that the IoD of a more important feature is weighted higher. While there could be a range of similar metrics that can be defined, we choose the most intuitive one 
to use in this paper. Since our main purpose is to complement Kendall's W in some corner cases and double check our experimental conclusions, rather than to justify any particular metric.


An example explanation used to illustrate the calculation of \eqref{eq_incon_def} is shown in Tab.~\ref{tab_example_incon_def}. First, the normalised coefficient vectors of the 3 repeat explanations\footnote{We set $k=3$ here for illustration, while generally $k$ should be large enough to let both the $f_i^{(k)}$ and $g_i^{(k)}$ converge for more accurate measurements.} are (presented in row)
\begin{align}
& [0.799, -0.599, 0.044,  0.036]\nonumber\\
&[ 0.726, -0.685, 0.044,  0.036] \nonumber\\ 
& [ 0.784,-0.619, 0.041,0.037]\nonumber
\end{align}
Thus,
\begin{align}
&\mathbb{E}(g_1^{(3)})=\frac{|0.799|+|0.726|+|0.784|}{3}=0.770\nonumber\\ &\mathbb{E}(g_2^{(3)})=\frac{|-0.599|+|-0.685|+|-0.619|}{3}=0.634\nonumber\\ &\mathbb{E}(g_3^{(3)})=\frac{|0.044|+|0.036|+|0.037|}{3}=0.039\nonumber\\ 
&\mathbb{E}(g_4^{(3)})=\frac{|0.036|+|0.044|+|0.041|}{3}=0.040\nonumber
\end{align}
Also,
\begin{align}
&\mathit{IoD}(f_1^{(3)})=\mathit{Var}([1,1,1])/\mathbb{E}([1,1,1])=0\nonumber\\ &\mathit{IoD}(f_2^{(3)})=\mathit{Var}([2,2,2])/\mathbb{E}([2,2,2])=0\nonumber\\ &\mathit{IoD}(f_3^{(3)})=\mathit{Var}([3,4,4])/\mathbb{E}([3,4,4])=0.09 \nonumber\\ &\mathit{IoD}(f_4^{(3)})=\nonumber\mathit{Var}([4,3,3])/\mathbb{E}([4,3,3])=0.1
\end{align}
Finally we get 0.00506 as the inconsistency via \eqref{eq_incon_def}.

\begin{table}[h]
\centering
\caption{An illustrative example of 3 repeated explanations on an instance with 4 features.}
\begin{tabular}{|c|c|c|c|c|c|c|}
\hline & \multicolumn{2}{c|}{exp.~1}                                           & \multicolumn{2}{c|}{exp.~2}                                           & \multicolumn{2}{c|}{exp.~3}                                           \\ \hline
\begin{tabular}[c]{@{}c@{}} feature\\rankings\end{tabular} & \begin{tabular}[c]{@{}c@{}} fea.\\ ID\end{tabular} & coef. & \begin{tabular}[c]{@{}c@{}}fea.\\ ID\end{tabular} & coef. & \begin{tabular}[c]{@{}c@{}}fea.\\ ID\end{tabular} & coef. \\ \hline
1st  & 1    & 0.2    & 1     & 0.18 & 1     & 0.19       \\ 
\hline
2nd & 2    & -0.15    & 2    & -0.17   & 2 &  -0.15      \\ \hline
3rd  & 3    & 0.011    & 4     & 0.011  & 4  & 0.01    \\ \hline
4th  & 4     & 0.009    & 3    & 0.009  & 3     & 0.009     \\ \hline
\end{tabular}
\label{tab_example_incon_def}
\end{table}
Fig.s~\ref{fig_incon} and \ref{fig_incon_kenw_image} show more experimental results of both Kendall's W and the inconsistency metric Eq.~\eqref{eq_incon_def} based on more instances, e.g., images, confirming our conclusions on RQ1. As expected, the plots of metric Eq.~\eqref{eq_incon_def} can be interpreted similarly as Kendall's W. But Kendall's W only considers the (discrete) ranks of each feature and treats all features equally. In contrast, our metric Eq.~\eqref{eq_incon_def} \textit{weights} the IoD of the discrete ranks of each feature by its (continuous) importance. With the extra information considered (compared to Kendall's W), our metric Eq.~\eqref{eq_incon_def} thus complements Kendall's W in terms of: (i) discriminating explanations with the same ranks of features but the importance vectors are different; (ii) minimising the effect from the normal fluctuation of the ranks of less-important/irrelevant features.

\section{Preliminaries on LIME}
\label{sec_app_preliminaries}

LIME \citep{ribeiro_why_2016} implements a local surrogate model that is used to explain individual predictions of black-box AI models. The intuition behind LIME is as follows. For a given black-box AI model (\textbf{model-agnostic}), we may probe it as many times as possible by perturbing some features (e.g., hiding superpixels of an image) of the input instance of interest (\textbf{locally}) and see how the prediction changes. Such changes in the predictions, as we vary the features of the instance, can help us understand why the AI model made a certain prediction over the instance.
Then a new dataset consisting of the perturbed inputs and the corresponding predictions made by the black-box AI model can be generated, upon which LIME trains a surrogate model that is \textbf{interpretable} to humans (e.g., linear regressors, cf. \citep[Chpt.~4]{molnar_interpretable_2020} for more). The training of the interpretable surrogate model is weighted by the proximity of the perturbed samples to the original instance. 

To be exact, the basic steps of how LIME works are:

\textit{1)} Choose an instance $\pmb{x}$ to interpret (e.g., an image or a row from a tabular dataset) in which there are $m$ features\footnote{LIME allows the selection of top few features by K-LASSO, while for brevity, we use all features when $n>m$ and top-5 features when $n \le m$.} (e.g., the superpixels for images or columns for tabular data). 

\textit{2)} Do perturbation on $\pmb{x}$ (e.g., switch on/off image superpixels) and generate a new input set $\pmb{X}=\{\pmb{x_1},...,\pmb{x_n}\}$ with $n$ samples, i.e. $\pmb{X}=(x_{ij}) \in \mathbb{R}^{n\times m}$. 

\textit{3)} Probe the black-box AI model with $\pmb{X}$ and record the predictions as a column vector $\pmb{y}=[y_1,...,y_n]^T$.

\textit{4)} Weight the $n$ perturbed samples in $\pmb{X}$ according to their proximity to the original instance $\pmb{x}$. Say the weights calculated by some kernel function (by default, an exponential kernel defined on some kernel width) are $\{w_1,...,w_n\}$, and denote the new weighted dataset\footnote{Depends on the specific implementation, e.g., by Python scikit-learn-0.22.1, this is equivalent to train a linear regressor on a scaled dataset $(\mathit{diag}(\sqrt{w_
		1},\dots,\sqrt{w_n})\pmb{X},\mathit{diag}(\sqrt{w_1},\dots,\sqrt{w_n})\pmb{y})$.} as $(\pmb{X}',\pmb{y}')$.


\textit{5)} On the dataset $(\pmb{X}',\pmb{y}')$, train a linear regressor
\begin{equation}
	\label{eq_linear_reg}
	\pmb{y}'=\pmb{X}'\pmb{\beta}+\epsilon
\end{equation}
where $\pmb{\beta}$ and $\epsilon$ are the coefficients and Gaussian white noise. 

\textit{6)} Then, the \textit{absolute values} of the coefficient vector $\pmb{\beta}$ 
represent the importance of the $m$ features, based on which rankings can be done. By default, LIME \footnote{LIME's Python implementation at \url{https://github.com/marcotcr/lime/}.} 
uses the Ridge regressor with weighted samples: 
\begin{equation}
	\pmb{\beta}=(\pmb{X}^{T}\pmb{W}\pmb{X}+r\pmb{I})^{-1} \pmb{X}^T\pmb{W}\pmb{y}
\end{equation}
where $r$ is the Ridge regularisation parameter, and $\pmb{W}=\mathit{diag(w_1,...,w_n)}$ is a diagonal matrix with diagonal elements equal to those $w_i$'s. It becomes the weighted ordinary least squares estimate $\pmb{\beta}=(\pmb{X}^{T}\pmb{W}\pmb{X})^{-1} \pmb{X}^T\pmb{W}\pmb{y}$ when $r=0$.

\end{document}